\documentclass{article}

    \PassOptionsToPackage{compress}{natbib}

    \usepackage[final]{neurips_2024}

\usepackage[utf8]{inputenc} %
\usepackage[T1]{fontenc}    %
\usepackage{hyperref}       %
\usepackage{url}            %
\usepackage{booktabs}       %
\usepackage{amsfonts}       %
\usepackage{nicefrac}       %
\usepackage{microtype}      %
\usepackage{xcolor}         %
\usepackage{graphicx}       %
\usepackage{subcaption}     %
\usepackage{tikz}           %
\usepackage{multicol}       %
\usepackage{multirow}       %
\usepackage{xspace}         %
\usepackage{amsmath}        %

\newcommand{\method}{\textsc{AOC}\xspace}

\title{Better Prompt Compression Without Multi-Layer Perceptrons}

\author{%
  Edouardo Honig\textsuperscript{1} \\
  \texttt{e.honig@ucla.edu} \\
  \And
  Andrew Lizarraga\textsuperscript{1} \\
  \texttt{andrewlizarraga@g.ucla.edu} \\
  \And
  Zijun Frank Zhang\textsuperscript{2} \\
  \texttt{fzhang@natera.com} \\
  \And
  Ying Nian Wu\textsuperscript{1} \\
  \texttt{ywu@stat.ucla.edu} \\
  \AND \vspace{-5mm}\\
  \textsuperscript{1}University of California, Los Angeles: Department of Statistics \& Data Science \\
  \textsuperscript{2}Natera \\
}

\begin{document}

\maketitle

\begin{abstract}
Prompt compression is a promising approach to speeding up language model inference without altering the generative model. 
Prior works compress prompts into smaller sequences of learned tokens using an encoder that is trained as a Low-Rank Adaptation (LoRA) of the inference language model. 
However, we show that the encoder does not need to keep the original language model's architecture to achieve useful compression. 
We introduce the Attention-Only Compressor (\method), which learns a prompt compression encoder after removing the multi-layer perceptron (MLP) layers in the Transformer blocks of a language model, resulting in an encoder with roughly 67\% less parameters compared to the original model. 
Intriguingly we find that, across a range of compression ratios up to $480\times$, \method can better regenerate prompts and outperform a baseline compression encoder that is a LoRA of the inference language model without removing MLP layers. 
These results demonstrate that the architecture of prompt compression encoders does not need to be identical to that of the original decoder language model, paving the way for further research into architectures and approaches for prompt compression. 
\end{abstract}

\vspace{-2mm}
\section{Introduction}
\vspace{-2mm}
Large language models (LLMs) display incredible usefulness across many natural language tasks, and generally have increased utility with increasingly long and complex prompts \citep{agarwal2024many, bertsch2024context}. 
The downside of lengthier prompts is increased computational load and response time, motivating research into compressing prompts into a smaller number of tokens, known as prompt compression. 

While some methods focus on compressing prompts by pruning information in the prompt/text space \citep{li2023compressing, jiang2023llmlingua, jiang2023longllmlingua}, one can also consider compressing prompts into a lower dimensional latent space \citep{wingate2022prompt, mu2023learning, chevalier2023adapting}. 
The In-context Autoencoder (ICAE) \citep{ge2024context} exemplifies this approach by training a LLM encoder to compress prompts into a shorter sequence of learned memory tokens and uses a learned \textbf{[AE]} autoencoder token for decoding the original prompt. 
This latent representation retains the information of the prompt and is used with the original frozen (meaning not further trained) LLM decoder to reduce the number of tokens at inference time. 
500xCompressor \citep{li2024xcompressor} works similarly, but compresses prompts into neural attention \citep{vaswani2017attention} key-value pairs instead of explicit tokens, and uses a pretrained \textbf{[BOS]} token instead of a learned \textbf{[AE]} token. 
Notably, both ICAE and 500xCompresson use Low-Rank Adaptation (LoRA) \citep{hu2021lora} to train encoders from the frozen decoder LLM used for inference, which requires more computational resources to perform compression than may be necessary. 

We demonstrate that using the entire decoder LLM as an encoder is unnecessary and introduce an alternative in the Attention-Only Compressor (\method). 
Instead of learning the encoder as a LoRA of the decoder LLM, we first remove the multi-layer perceptron (MLP) layers before training the entire encoder. 
By removing MLPs, \method's prompt compression encoder has roughly 67\% less parameters compared to previous methods' encoders, while improving or maintaining similar compression ability. 
These results emphasize that prompt compression encoders do not need identical architecture to their decoders and that there exist compression models that with higher performance and lower inference-time computational requirements compared to recent approaches using frozen-LLM-based compressors.

Our contributions can be summarized as follows:
\begin{itemize}
    \item We introduce the Attention-Only Compressor (\method), a novel prompt compression encoder that removes the MLP layers from a LLM, resulting in an encoder that performs comparably to baseline compression encoders that are roughly three times larger. 
    \item Preliminary experimental results on regeneration demonstrate that compression encoders do not need architecture identical to their decoders, which motivates further research into more efficient compressors. 
    \item To further study compression encoders, we present examples of interpolating between the embeddings of two compressed prompts, showcasing a novel classifier-free approach to merging separate prompts and understanding the latent space of compressed prompts.
\end{itemize}

\vspace{-2mm}
\section{Methods}
\vspace{-2mm}
\textbf{Model}. 
Our proposed model consists of a learned prompt compression encoder $\mathbf{E}$ and a pretrained LLM decoder $\mathbf{D}$ that is always frozen throughout training and inference. 
The encoder is architecturally identical to the decoder as in 500xCompressor and ICAE, with the key exception that the MLP layers have been replaced with the identity operation within each block of the Transformer \citep{vaswani2017attention}:
\vspace{-2mm}
\begin{align}
    h_\ell &= {\rm LN_{pre}}(h_{\ell-1}) & h_\ell &= {\rm LN_{pre}}(h_{\ell-1}) \\
    h_\ell &= {\rm MHA}(h_\ell) + h_\ell &     h_\ell &= {\rm MHA}(h_\ell) + h_\ell \\
    h_\ell &= {\color{red}{\rm MLP}({\rm LN_{post}}(h_\ell))} + h_\ell &    h_\ell &= {\color{blue}{\rm LN_{post}}(h_\ell)} + h_\ell
\end{align}
$h_{\ell-1}$ denotes the input hidden state to the $\ell$th Transformer block, $\rm LN_{pre}$ and $\rm LN_{post}$ are layer norms \citep{ba2016layer}, and $\rm MHA$ denotes multi-headed attention \citep{vaswani2017attention}. 

Let the input for the encoder be represented by the concatenation of $n$ prompt tokens $\mathbf{X}_n = (x_1, \dots, x_n)$ with the encoder's $m$ learned memory tokens $\mathbf{Y}_m = (y_1, \dots, y_m)$. 
$\mathbf{Z}=\mathbf{E}([\mathbf{X}_n,\mathbf{Y}_m])$ is the latent representation from the encoder output. 
For 500xCompressor, $\mathbf{Z}=\{\mathbf{KV}(h_{\ell}^{\mathbf{Y}_m}) \forall \ell \}$: the encoder's per-layer attention key-value pairs corresponding to $\mathbf{Y}_m$. 
The input to the decoder is $\mathbf{Z}$ concatenated with a regeneration token \textbf{[REGEN]}, which is used to regenerate $\mathbf{X}$ using the latent information from $\mathbf{E}$. 
For both 500xCompressor and \method \textbf{[REGEN]} is the \textbf{[BOS]} token. 
Therefore, the regeneration of $\mathbf{X}_n$ from the latent representation $\mathbf{Z}$ is given by
\begin{equation}\label{eqn:regen}
    \hat{\mathbf{X}}_n = \mathbf{D}\left(\left[\mathbf{Z}, \mathbf{[REGEN]}\right]\right) = \mathbf{D}\left(\left[\mathbf{E}([\mathbf{X}_n,\mathbf{Y}_m]), \mathbf{[BOS]}\right]\right)
\end{equation}
The standard cross-entropy loss between the decoder logits and the input $\mathbf{X}$ is used to train the encoder via backpropagation \citep{lecun1989backpropagation}. 
For all experiments, we use Llama 3.2 1B Instruct \citep{meta2024llama} as the pretrained LLM in bfloat16 \citep{cloud2019bfloat16} precision, AdamW \citep{kingma2014adam} with a 300-step warmup to a learning rate $2\times10^{-4}$ as the optimizer in PyTorch \citep{paszke2019pytorch} conducting training using Transformers \citep{wolf2020transformers} on a single NVIDIA A6000 GPU. 
LoRAs are trained on the queries, keys, values, and output projections in the multi-headed attention components for 500xCompressor and LoRA ablations on \method. 

\textbf{LoRA Ablations}. 
Due to the lower number of total parameters in the encoder for \method, we perform full training instead of LoRA to learn a strong prompt compressor. 
However, this causes the total number of parameters in memory at both training and inference time to be slightly larger with \method compared to the baseline 500xCompressor which use a LoRA of the decoder LLM. 
This trade-off of increased memory for decreased compression time motivates ablations on learning the \method encoder using LoRA (LoRA-\method) instead of training the entire encoder. 

\textbf{Compressed Prompt Interpolation}. 
The latent information $\mathbf{Z}$ from compressing a prompt has not been extensively studied beyond classifier-guided generation by \cite{wingate2022prompt}. 
As an initial step toward better understanding the compressed forms of prompts, we conduct linear interpolations between compressed prompts and qualitatively inspect the intermediary output. The interpolation between $\mathbf{Z}_0$ and $\mathbf{Z}_1$ with a given weight $w$ is given by:
\begin{equation}
    \mathbf{Z}_{\rm interp} = \mathbf{Z}_0 + w (\mathbf{Z}_1 - \mathbf{Z}_0)
\end{equation}
\textbf{Data}. 
Experiments are performed using random samples from the arXiv dataset \citep{arxiv_org_submitters_2024}. 
\method is trained on 300,000 abstracts from the arXiv dataset first submitted before July 1, 2023 and validated on 3,000 abstracts first submitted after January 4, 2024. 
Final evaluations were conducted on a held-out test set of 3,000 abstracts from after January 4, 2024. 
These dates were chosen based on the Llama 3.2 training cutoff of December 2023, and are identical to the cutoffs presented in \citep{li2024xcompressor}. 
The amount of training data was determined while accounting for limited computational resources. 

\textbf{Metrics}. 
We evaluate \method on text regeneration as performed using \autoref{eqn:regen}. 
Following \citep{ge2024context}, we report the Bilingual Evaluation Understudy (BLEU) \citep{papineni2002bleu} and Exact-Match (EM) scores. 
Notably, the EM metric defined by \citep{ge2024context} is the proportion of identical prefix length to total target length.
Given a regenerated sequence of length $n'$, this proportional EM metric is defined as:
\begin{equation}\label{eqn:em-metric}
    {\rm EM}(\mathbf{X}_n,\hat{\mathbf{X}}_{n'}) = \frac{1}{n}\sum_{i=1}^{n} \mathbf{1}_{\mathbf{X}_i = \hat{\mathbf{X}}_{i}}(\mathbf{X}_i, \hat{\mathbf{X}}_i)
\end{equation} 
In contrast, the EM metric defined by \citep{li2024xcompressor} is a binary metric equal to $1$ when the regeneration $\hat{\mathbf{X}}_{n'}$ is identical to $\mathbf{X}_n$ and $0$ otherwise, introducing a discrepancy in notation. 
We report the EM metric as defined in \autoref{eqn:em-metric} since it is more informative. 
Additionally, we report the Recall-Oriented Understudy for Gisting Evaluation Longest Common Subsequence (ROUGE-L) \citep{lin2004rouge} F1 scores which evaluate overall sequence similarity, following \citep{li2024xcompressor}.

\vspace{-2mm}
\section{Results}
\vspace{-2mm}

\textbf{Baseline Comparison}. 
To demonstrate the benefits of \method, we compare to 500xCompressor with a variety of input prompt lengths $n \in \{96, 192, 288, 384, 480\}$ and number of memory tokens $m \in \{1, 4, 16\}$. 

\begin{table}[h]
  \caption{Evaluation results for models trained with $m=16$ memory tokens.}
  \label{tab:scores_m16}
  \centering
  \begin{tabular}{clccc}
    \toprule
    Prompt Length & Model & BLEU ($\uparrow$) & EM ($\uparrow$) & ROUGE-L F1 ($\uparrow$) \\
    \midrule
    \multirow{3}{*}{$n=96$} 
    & 500xCompressor & 0.981 & 0.740 & 0.990 \\
    & LoRA-\method   & 0.740 & 0.197 & 0.856 \\
    & \method        & \textbf{0.984} & \textbf{0.889} & \textbf{0.991} \\
    \midrule
    \multirow{3}{*}{$n=192$} 
    & 500xCompressor & 0.850 & 0.109 & 0.915 \\
    & LoRA-\method   & 0.284 & 0.065 & 0.510 \\
    & \method        & \textbf{0.868} & \textbf{0.454} & \textbf{0.924} \\
    \midrule
    \multirow{3}{*}{$n=288$} 
    & 500xCompressor & 0.685 & 0.130 & 0.816 \\
    & LoRA-\method   & 0.319 & 0.068 & 0.548 \\
    & \method        & \textbf{0.839} & \textbf{0.465} & \textbf{0.901} \\
    \midrule
    \multirow{3}{*}{$n=384$} 
    & 500xCompressor & 0.662 & 0.106 & 0.799 \\
    & LoRA-\method   & 0.255 & 0.068 & 0.478 \\
    & \method        & \textbf{0.801} & \textbf{0.386} & \textbf{0.880} \\
    \midrule
    \multirow{3}{*}{$n=480$} 
    & 500xCompressor & 0.588 & 0.082 & 0.746 \\
    & LoRA-\method   & 0.201 & 0.053 & 0.421 \\
    & \method        & \textbf{0.823} & \textbf{0.483} & \textbf{0.893} \\
    \bottomrule
  \end{tabular}
\end{table}

As seen in \autoref{tab:scores_m16} and \autoref{tab:scores_m4}, \method outperforms 500xCompressor across all prompt lengths with 4 or 16 memory tokens despite having 67\% less encoder parameters. 
Based on the results in \autoref{tab:scores_m16} we find that \method and 500xCompressor only perform similarly when restricted to a single memory token. 
The large variance in EM between models can be attributed to differences in early parts of the regeneration, as the EM metric is based on the proportion of identical prefix matching. 
Interestingly, LoRA-\method tends to perform worse than \method and the baseline 500xCompressor across all metrics, which suggests that the effectiveness of LoRA in Transformers relies in part on the frozen MLPs, in line with prior work on freezing Transformer components \cite{lu2022pretrained}. 

\begin{table}[h]
  \caption{Evaluation results for models trained with $m=4$ memory tokens.}
  \label{tab:scores_m4}
  \centering
  \begin{tabular}{clccc}
    \toprule
    Prompt Length & Model & BLEU ($\uparrow$) & EM ($\uparrow$) & ROUGE-L F1 ($\uparrow$) \\
    \midrule
    \multirow{3}{*}{$n=96$} 
    & 500xCompressor & 0.669 & 0.073 & 0.815 \\
    & LoRA-\method   & 0.342 & 0.069 & 0.599 \\
    & \method        & \textbf{0.711} & \textbf{0.221} & \textbf{0.843} \\
    \midrule
    \multirow{3}{*}{$n=192$} 
    & 500xCompressor & 0.302 & 0.015 & 0.561 \\
    & LoRA-\method   & 0.136 & 0.021 & 0.399 \\
    & \method        & \textbf{0.374} & \textbf{0.064} & \textbf{0.615} \\
    \midrule
    \multirow{3}{*}{$n=288$} 
    & 500xCompressor & 0.218 & 0.035 & 0.484 \\
    & LoRA-\method   & 0.126 & 0.019 & 0.387 \\
    & \method        & \textbf{0.339} & \textbf{0.056} & \textbf{0.578} \\
    \midrule
    \multirow{3}{*}{$n=384$} 
    & 500xCompressor & 0.236 & 0.013 & 0.507 \\
    & LoRA-\method   & 0.117 & 0.015 & 0.378 \\
    & \method        & \textbf{0.300} & \textbf{0.040} & \textbf{0.558} \\
    \midrule
    \multirow{3}{*}{$n=480$} 
    & 500xCompressor & 0.241 & 0.027 & 0.508 \\
    & LoRA-\method   & 0.068 & 0.011 & 0.288 \\
    & \method        & \textbf{0.343} & \textbf{0.058} & \textbf{0.587} \\
    \bottomrule
  \end{tabular}
\end{table}

It can be seen in \autoref{tab:scores_m1} that for $m=1$, \method performs on-par with 500xCompressor, although both display poor regeneration abilities for some of the largest compression ratios in our experiments. 
Upon inspection of the loss curves from training the $m=1$ models in \autoref{tab:scores_m1}, we discover that they are likely under-trained due to computational budget constraints. 
Based on these results, it appears that increasing the amount of memory tokens $m$ may allow for a smaller training data set. 

\begin{table}[h]
  \caption{Evaluation results for models trained with $m=1$ memory token.}
  \label{tab:scores_m1}
  \centering
  \begin{tabular}{clccc}
    \toprule
    Prompt Length & Model & BLEU ($\uparrow$) & EM ($\uparrow$) & ROUGE-L F1 ($\uparrow$) \\
    \midrule
    \multirow{3}{*}{$n=96$} 
    & 500xCompressor & 0.122 & 0.013 & \textbf{0.382} \\
    & LoRA-\method   & 0.092 & 0.022 & 0.355 \\
    & \method        & \textbf{0.129} & \textbf{0.037} & 0.369  \\
    \midrule
    \multirow{3}{*}{$n=192$} 
    & 500xCompressor & \textbf{0.102} & 0.015 & \textbf{0.352} \\
    & LoRA-\method   & 0.074 & 0.013 & 0.308 \\
    & \method        & 0.095 & \textbf{0.017} & 0.327 \\
    \midrule
    \multirow{3}{*}{$n=288$} 
    & 500xCompressor & \textbf{0.090} & 0.007 & \textbf{0.337} \\
    & LoRA-\method   & 0.061 & 0.009 & 0.278 \\
    & \method        & 0.089 & \textbf{0.016} & 0.317 \\
    \midrule
    \multirow{3}{*}{$n=384$} 
    & 500xCompressor & 0.089 & 0.009 & \textbf{0.337} \\
    & LoRA-\method   & 0.068 & 0.010 & 0.302 \\
    & \method        & \textbf{0.094} & \textbf{0.019} & 0.330 \\
    \midrule
    \multirow{3}{*}{$n=480$} 
    & 500xCompressor & 0.094 & 0.004 & \textbf{0.355} \\
    & LoRA-\method   & 0.056 & 0.008 & 0.273 \\
    & \method        & \textbf{0.097} & \textbf{0.015} & 0.341 \\
    \bottomrule
  \end{tabular}
\end{table}

\textbf{Latent Space Inspection}.
In \autoref{tab:interpolation} we show the result of linearly interpolating between the compressed information $\mathbf{Z}$ from the prompt $p_0=$\texttt{"We present an awesome new idea."} and the prompt $p_1=$\texttt{"Large planets may have many moons."} for \method, color-coding by similarity to $p_0$ or $p_1$. 
As can be observed, the interpolation of the two latent representations results in a regenerated mixture of prompts, such as when the interpolation weight $w=0.5$ \texttt{planet} which is more closely related to \texttt{planets} from $p_1$ than \texttt{idea} from $p_0$. 
For $w=0.53$, \texttt{many moons} from $p_1$ appears in a regeneration that shares the same prefix as $p_0$. 
Similarly, for interpolation weights $w=0.55$ and $w=0.6$, \texttt{amazing} and \texttt{wonderful}, which are more closely related to \texttt{awesome} from $p_0$, appear in a regeneration almost identical to $p_1$ with the same two-word prefix. 
We also note that \autoref{tab:interpolation} shows both $p_0$ and $p_1$ were perfectly regenerated from their unaltered compressed states with zero information loss. 

\begin{table}[h]
  \caption{Regeneration of linearly interpolated latent information.}
  \label{tab:interpolation}
  \centering
    \begin{tabular}{cl}
      \toprule
      Interpolation Weight & Regeneration \\
      \midrule
      $w=0.00$ & \textcolor{blue}{We present an awesome new idea.}\\
      $w=0.40$ & \textcolor{blue}{We present an} \textcolor{teal}{amazing} \textcolor{blue}{new idea.}\\
      $w=0.50$ & \textcolor{blue}{We present an} \textcolor{teal}{amazing} \textcolor{blue}{new} \textcolor{teal}{planet.}\\
      $w=0.53$ & \textcolor{blue}{We present an} \textcolor{teal}{amazing} \textcolor{orange}{many moons.}\\
      $w=0.55$ & \textcolor{orange}{Large planets have many} \textcolor{blue}{amazing.}\\
      $w=0.60$ & \textcolor{orange}{Large planets have many} \textcolor{blue}{wonderful.}\\
      $w=1.00$ & \textcolor{orange}{Large planets may have many moons.}\\
      \bottomrule
    \end{tabular}
  \vspace{2mm}
\end{table}

\vspace{-2mm}
\section{Conclusion}
\vspace{-2mm}
We introduce \method, a prompt compression encoder using only attention layers from a decoder LLM that demonstrably achieves comparable or better compression to LoRA baselines with identical architecture to the decoder LLM. 
Experiments show that the memory tokens learned with \method can encode similar amounts of information to baselines with $3 \times$ the amount of parameters. 
In future work, we hope to further explore encoder architectures, as our results indicate that a prompt compression encoder need not have the same architecture as the decoder LLM. 
Additionally, we seek to better understand the latent space formed by compressed prompts and extend the use of compressed prompts beyond the interpolation example presented in this work. 
While this work was performed with limited computational resources, we aim to study more diverse and larger datasets, model architectures, and compression ratios in the future.

\bibliography{0_main}

\newpage
\clearpage

\end{document}